\documentclass[final]{cvpr}

\usepackage{times}
\usepackage{epsfig}
\usepackage{graphicx}
\usepackage{amsmath}
\usepackage{amssymb}
\usepackage{multirow}
\usepackage{verbatim}
\usepackage{mathrsfs}
\usepackage{bbm}
\usepackage{subfigure}  
\usepackage[noend]{algpseudocode}
\usepackage{algorithmicx,algorithm}

\usepackage[pagebackref=true,breaklinks=true,colorlinks,bookmarks=false]{hyperref}

\def\cvprPaperID{233} 
\def\confYear{CVPR 2021}
\newcommand{\tabincell}[2]{\begin{tabular}{@{}#1@{}}#2\end{tabular}}

\begin{document}

\title{Neural Feature Search for RGB-Infrared Person Re-Identification}

\author{Yehansen Chen$^{1}$\footnotemark[1], Lin Wan$^{1}$\footnotemark[1], Zhihang Li$^{2}$\footnotemark[2], Qianyan Jing$^{1}$, Zongyuan Sun$^{1}$\\
$^{1}$School of Geography and Information Engineering, China University of Geosciences, Wuhan, China\\
$^{2}$School of Artificial Intelligence, University of Chinese Academy of Sciences, Beijing, China\\
}
\maketitle
\pagestyle{empty}
\thispagestyle{empty}
\renewcommand{\thefootnote}{$*$}
\footnotetext[1]{Equally-contributed first authors}
\renewcommand{\thefootnote}{$\dag$}
\footnotetext[2]{Corresponding author}

\begin{abstract}
   RGB-Infrared person re-identification (RGB-IR ReID) is a challenging cross-modality retrieval problem, which aims at matching the person-of-interest over visible and infrared camera views. Most existing works achieve performance gains through manually-designed feature selection modules, which often require significant domain knowledge and rich experience. In this paper, we study a general paradigm, termed Neural Feature Search (NFS), to automate the process of feature selection. Specifically, NFS combines a dual-level feature search space and a differentiable search strategy to jointly select identity-related cues in coarse-grained channels and fine-grained spatial pixels. This combination allows NFS to adaptively filter background noises and concentrate on informative parts of human bodies in a data-driven manner. Moreover, a cross-modality contrastive optimization scheme further guides NFS to search features that can minimize modality discrepancy whilst maximizing inter-class distance. Extensive experiments on mainstream benchmarks demonstrate that our method outperforms state-of-the-arts, especially achieving better performance on the RegDB dataset with significant improvement of 11.20\% and 8.64\% in Rank-1 and mAP, respectively.
\end{abstract}

\section{Introduction}
Person re-identification (ReID) aims to match the person-of-interest over non-overlapping camera views \cite{wang2014person, ye2020deep, tang2019unsupervised, zou2020joint, wu2020tracklet}, serving as a central part of intelligent video surveillance systems. Currently, most conventional ReID methods concentrate efforts on visible images-based cross-view matching tasks \cite{liu2018pose, hou2019interaction, tian2018eliminating, shen2018person, miao2019pose}, which cannot adapt well to illumination variations in real-world scenarios (e.g., low lighting environments at nighttime). Motivated by this challenge, associating RGB and infrared (IR) pedestrian images captured by dual-mode cameras for cross-modality image retrieval, a.k.a. RGB-IR ReID, has drawn much attention in vision community \cite{wu2017rgb, ye2018hierarchical, wang2019rgb, wang2020cross, ye2020dynamic}.

\begin{figure}[t]
\begin{center}
   \includegraphics[width=1\linewidth]{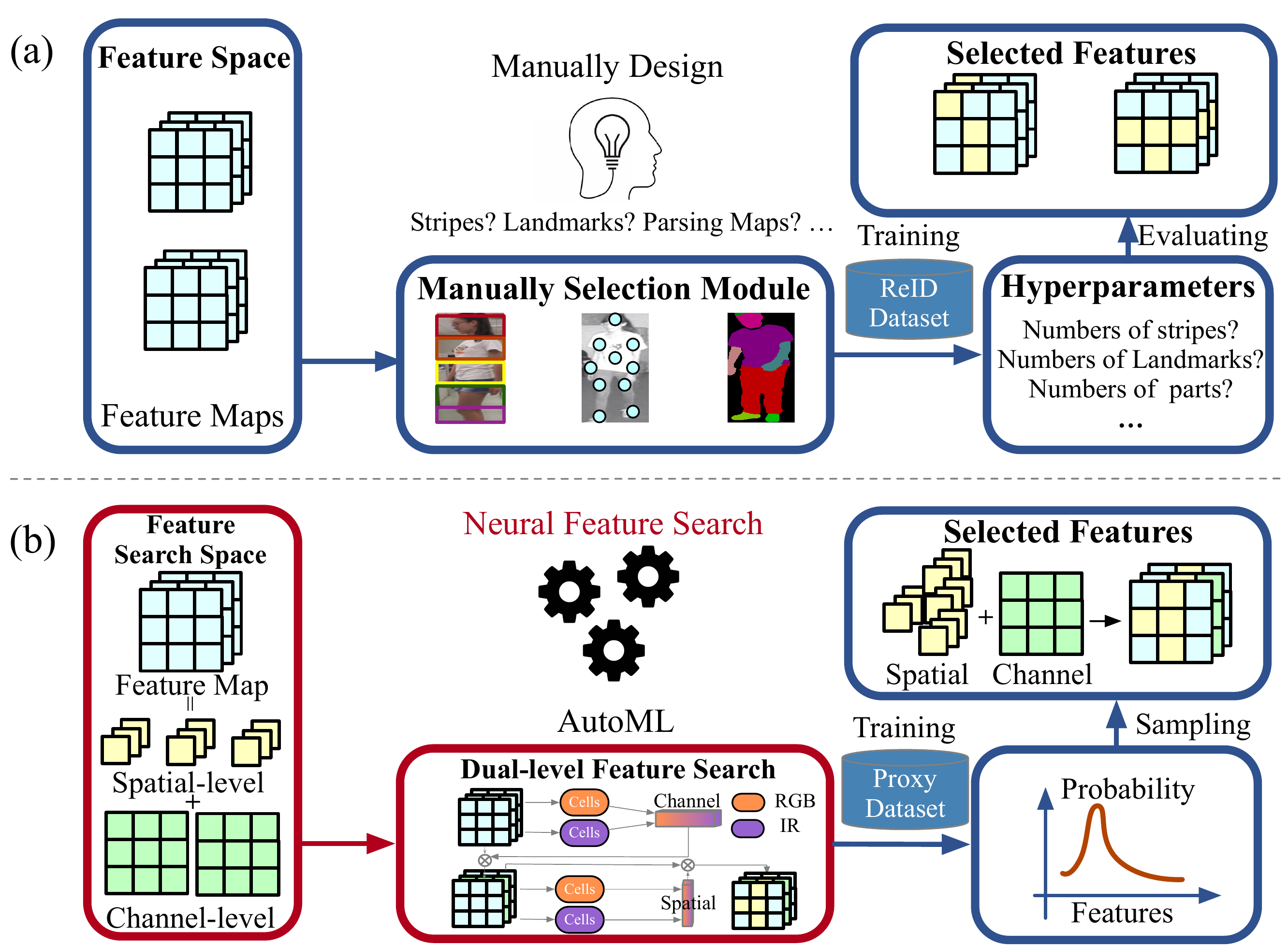}
\end{center}
   \caption{Comparison of hand-crafted and automated feature selection strategies. (a) Manually designing task-specific modules to select identity-related features.
   (b) NFS automatically derives the optimal feature subset from a dual-level feature search space.}
   \vspace{-0.5cm} 
\label{fig:problem}
\end{figure}

Due to intrinsically different imaging mechanisms, RGB-IR ReID suffers from undesired visual discrepancy between visible and infrared images, which makes appearance cues such as colors and textures unreliable or even misleading for the matching task \cite{lu2020cross, wang2020cross, wu2017rgb}. Moreover, such modality divergence also exacerbates the already large intra-class variations caused by diverse camera viewpoints, person poses, partial occlusions, and background clutter \cite{wu2020rgb, hao2019hsme, li2020infrared}, making it even harder to align images of the same identity. In an effort to minimize the modality gap, cross-modality image synthesis methods \cite{wang2019rgb, wang2020cross, li2020infrared} typically leverage generative adversarial networks (GANs) to transfer stylistic properties between modalities to synthesize fake RGB/IR images. But it is non-trivial to preserve identity information for generated RGB images due to lack of color information in their IR counterparts \cite{gong2020feature}. Another line of shared feature learning approaches \cite{ye2018hierarchical, ye2018visible, ye2020dynamic, ye2020cross} utilize convolutional neural networks (CNNs) to perform cross-modality feature alignment. One representative model-of-choice is the two-stream network \cite{ye2020cross, ye2020visible, wei2020co}, which includes modality-specific shallow layers and shared deeper layers to learn a common feature space \cite{wu2017rgb}. On the strength of two-stream structures, several studies exploit ReID discriminative constrains, e.g., triplet loss \cite{ye2020deep, ye2019bi, ye2020cross} or ranking loss \cite{ye2018visible, hao2019hsme}, to supervise the network to mine identity-related cues. They are all committed to learning a better distance metric that enhances the performance of similarity-based retrieval and have achieved significant success in recent years \cite{ye2020dynamic}.

To our understanding, whether image synthesis approaches or shared feature learning techniques, the crux of ReID solutions is always to find sufficiently high-quality discriminative features for matching and retrieval. To achieve this goal, state-of-the-art methods typically introduce partition stripes \cite{ye2020dynamic}, human landmarks \cite{wang2020high}, parsing maps \cite{kalayeh2018human}, and body contour sketches \cite{yang2019person} to discourage irrelevant features whilst preserving the identity-related ones (Fig. \ref{fig:problem}(a)). However, it is really tough and time-consuming to manually design a \textit{one-fit-all} feature selection module against all sorts of intra- and inter-modality variations, leading to unsatisfactory performance of human-guided feature selection mechanisms. Driven by the above observations, a question arises: \textit{Is there a data-driven feature selection manner without much requirement for human interference?} Recent advances on automated machine learning (AutoML) \cite{he2021automl} may provide a positive answer. Using Neural Architecture Search (NAS) \cite{zoph2016neural}, Quan \textit{et al.} \cite{quan2019auto} automatically generate fast and effective CNNs whose performance is on par with hand-crafted architectures in single-modality ReID tasks. This progress inspires the idea of neuron-powered automatic feature selection discussed in this paper.

With the idea in mind, we investigate RGB-IR ReID from a \textit{one-fit-all} feature search perspective. Different from existing manually-designed \cite{ye2020dynamic, lu2020cross, choi2020hi} or NAS-generated network structures \cite{quan2019auto}, our goal is to pursue better ReID performance by discovering discriminative features with data-driven search neurons. To this end, we cast feature selection as a bilevel optimization problem \cite{liu2018darts} (i.e., deriving the optimal feature subset from the best feature learning results) and propose a novel paradigm, \textit{Neural Feature Search} (NFS, Fig. \ref{fig:problem}(b)). Starting from the hierarchical feature extraction mechanism of CNNs \cite{perez2019mfas}, NFS includes a dual-level feature search space where each feature map is decomposed in terms of pixel and channel dimensions. This allows feature selection operations to be jointly performed in a mutually reinforcing manner—channel-level search identifies relevant response maps from the global view while pixel-level search scans every spatial position to selectively process local part features of a person. To improve the search efficiency, we utilize reparameterization tricks \cite{loaiza2019continuous} to relax the search space to be continuous, which makes the optimization of search neurons compatible with stochastic gradient descent (SGD). Considering the inherent modality discrepancy issue of RGB-IR ReID, a cross-modality contrastive optimization scheme is further introduced as a supervision signal that discourages irrelevant features whilst encouraging modality-invariant cues during the searching process. Extensive experiments show that NFS significantly outperforms the state-of-the-arts by 12.01\% and 11.20\% gains of Rank-1 accuracy on SYSU-MM01 (\textit{multi-shot, all search} mode) and RegDB (\textit{visible-to-infrared} mode) benchmarks, respectively. To summarize, this paper brings three main contributions:

\begin{itemize}
    \item We propose an AutoML-powered neural feature search method for RGB-IR person re-identification, which automates the feature selection process to substantially improve the matching accuracy with less human efforts. To our best knowledge, this is one of the first attempts to utilize automatic feature selection techniques for cross-modality ReID.
    
    \item We introduce a novel feature search space allowing both spatial and channel-wise feature selection. Based on this search space, we present an efficient feature search algorithm embedded with a cross-modality contrastive optimization mechanism, effectively tackling the modality discrepancy in RGB-IR ReID.
    
    \item Extensive experiments on two mainstream RGB-IR ReID benchmarks demonstrate the superiority of NFS compared with previous state-of-the-arts. 
\end{itemize}

\section{Related Work}
\textbf{RGB-based Person ReID.} RGB-based person ReID studies mainly focus on handling intra-class variations of pose \cite{liu2018pose}, scale \cite{hou2019interaction}, and background clutter \cite{tian2018eliminating} presented in visible images. Nowadays, substantial research efforts \cite{hou2019interaction, wang2019color, liu2019adaptive, ding2020multi, yan2020learning} have been devoted to deep learning-based ReID for more effective feature learning and alignment. For example, graph neural networks \cite{shen2018person, miao2019pose} make full use of relationships between global embedding vectors to map images into a discriminative feature space. Self-attention based methods \cite{vaswani2017attention, luo2019spectral} explore pixel similarities to let the network concentrate on informative biometrics such as face against the background clutter. Apart from global feature representation learning, several local feature learning approaches \cite{wang2020high, zheng2019pose} also employ pretrained pose estimation models to decompose the human body into landmarks or parsing maps and perform fine-grained feature alignment over pose changes and occlusions. Although having achieved considerable success in reducing intra-class variations, most existing single-modality ReID methods are ill-suited for cross-modality image retrieval in poor lighting environments \cite{ye2020visible, ye2019bi, ye2020cross}.

\begin{figure*}[t]
\begin{center}
   \includegraphics[width=0.75\linewidth]{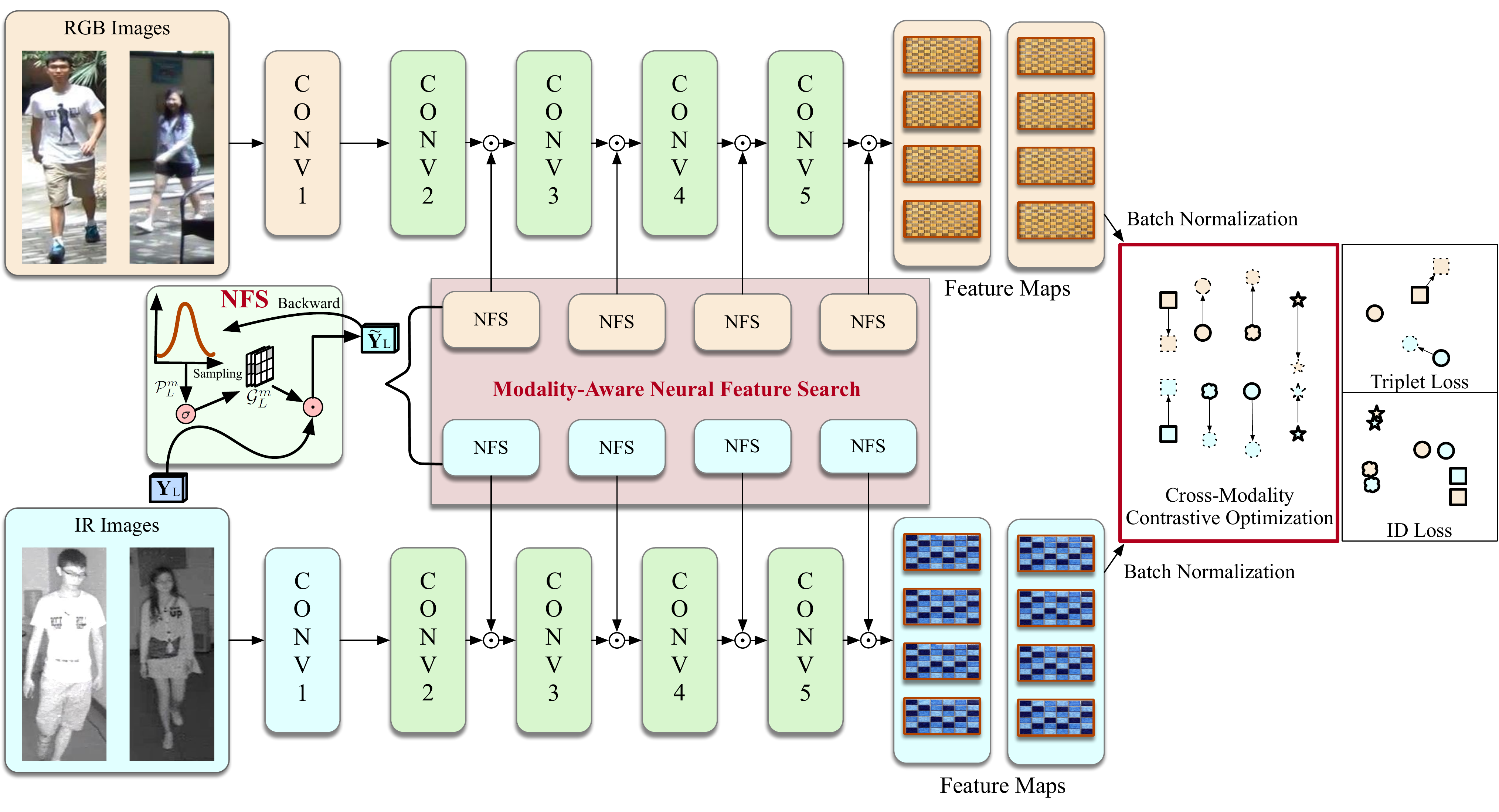}
\end{center}
   \caption{An overview of our NFS paradigm. It combines modality-aware search cells and cross-modality contrastive optimization mechanism to conduct automated feature selection on two-stream CNNs based feature space. Each learnable search cell is jointly optimized with network parameters to derive the optimal feature subset in every shared block. The cross-modality contrastive optimization mechanism further enables NFS to search modality-invariant features that can minimize modality discrepancy while maximizing inter-class distance.}
\label{fig:pipeline}
\end{figure*}

\textbf{RGB-Infrared Person ReID.} In addition to intra-class variations, RGB-IR ReID also considers the modality discrepancy issue caused by different wavelength ranges of visible and infrared cameras \cite{wang2020cross}. Current RGB-IR ReID researches mainly resort to either GAN-based \cite{wang2019rgb, wang2020cross, dai2018cross, choi2020hi, zhang2020rgb, wang2019learning} or shared feature learning approaches \cite{ye2018visible, ye2018hierarchical, ye2020dynamic, wu2017rgb, lu2020cross} to handle both intra- and inter-modality variations.

For GAN-based approaches, early attempts \cite{dai2018cross} usually adopt adversarial training strategies to reduce the distribution divergence of modality-specific features. Along a somewhat different line, Wang \textit{et al.} \cite{wang2019rgb} leverage GANs to transfer stylistic properties of IR images to their RGB counterparts for jointly pixel and feature alignment. Several studies also apply pair-wise pixel alignment \cite{wang2020cross}, feature disentanglement \cite{choi2020hi}, or intermediate modality generation \cite{li2020infrared, wang2019learning} to further eliminate appearance differences across modalities. However, it is non-trivial to accurately choose the suitable target for style transfer \cite{lu2020cross}, which may lead to identity inconsistency during the complicated adversarial training process \cite{gong2020feature}. As for the shared feature learning category, Wu \textit{et al.} \cite{wu2017rgb} first contribute a large benchmark dataset (SYSU-MM01) and propose a one-stream zero-padding network for RGB-IR image matching. Nowadays, two-stream CNNs based methods are dominating the cross-modality person ReID community. For instance, some recent studies extend two-stream CNNs with deep metric learning \cite{ye2018hierarchical, ye2018visible, hao2019hsme, ye2020cross, ye2019bi, ye2020visible, tekeli2019distance} or the attention mechanism \cite{ye2020dynamic, wei2020co} to learn modality-sharable representations against both modality discrepancy and high sample noises. Several works \cite{feng2019learning, lu2020cross} also employ modality-specific or modality-aware learning avenues to perform cross-modality identity recognition at the classifier level.

\textbf{Neural Architecture Search.} Recent years have witnessed a growing body of Neural Architecture Search (NAS) researches \cite{elsken2018neural, li2020gp} that have achieved considerable success in various domains, e.g., image classification \cite{real2019regularized}, semantic segmentation \cite{chen2019fasterseg}, object detection \cite{liang2019computation}, ReID \cite{quan2019auto}, and multi-modal fusion \cite{perez2019mfas}, etc. Generally, NAS aims to automatically search optimal operations or topology of deep neural networks for specific learning tasks. They first construct a task-oriented search space that defines which architectures can be discovered in principle. Based on the search space, different search strategies, including reinforcement learning-based methods \cite{guo2020breaking, pham2018efficient}, evolutionary methods \cite{liu2017hierarchical, real2019regularized}, gradient-based ones \cite{liu2018darts, lian2019towards}, and Monte Carlo Tree Search (MCTS) approaches \cite{negrinho2017deeparchitect, wang2020learning}, are proposed and prove effective for improving both sample efficiency and model performance. Inspired by the basic idea of NAS, we present a \textit{one-fit-all} feature selection strategy for RGB-IR person ReID, revolutionizing manually-crafted feature selection components in the existing literature.
\section{Methodology}
Fig. \ref{fig:pipeline} presents an overview of our proposed method. On the basis of a two-stream network (Section \ref{sec:3.1}), NFS mainly includes a dual-level search space for spatial and channel-wise feature selection, and a differentiable feature search algorithm (Section \ref{sec:3.2}) governed by the cross-modality contrastive optimization scheme (Section \ref{sec:3.3}) to prune discriminative cues fast and accurately.

\subsection{Baseline RGB-IR Person ReID}
\label{sec:3.1}
We adopt the two-stream CNN employed in \cite{ye2018visible, ye2020deep, ye2020dynamic} as the baseline network\footnote{https://github.com/mangye16/Cross-Modal-Re-ID-baseline}. To capture modality-invariant information, parameters of the first convolutional block are independent for each modality, while the other layers are shared to learn discriminative features \cite{wu2017rgb}. After the last convolutional layer with global average pooling, a shared batch normalization layer is used to attain final representations for heterogeneous images. During the training phase, we aim to minimize the following baseline loss function: 
\begin{equation}
    \mathcal{L}_{b} = \mathcal{L}_{id} + \mathcal{L}_{tri},
\end{equation}
where $\mathcal{L}_{id}$ is the softmax cross-entropy loss and $\mathcal{L}_{tri}$ is the weighted regularization triplet (WRT) loss \cite{ye2020deep}.

\subsection{Modality-aware Neural Feature Search}
\label{sec:3.2}
Unlike NAS approaches searching optimal topology and operations for a top-performing architecture \cite{liu2018darts, wu2019fbnet, pham2018efficient}, NFS searches for identity-related features from a CNN-based feature space. In this paper, we cast the automatic feature search as a hyperparameter learning task, where the search hyperparameters and network weights are jointly optimized to derive the optimal discriminative feature subset. It can be formulated as a bilevel optimization problem \cite{liu2018darts}:
\begin{equation}
\label{eq:2}
\min  \limits_{y\in Y} \min \limits_{W} \mathcal{L} (y, W).
\end{equation}
Given output feature maps $Y$, we seek to discover a subset of features $y\in Y$, which can minimize the loss $\mathcal{L} (y, W)$ after optimizing network weights $W$. Here, we highlight two key points of solving the problem: a dual-level feature search space and an efficient search algorithm.

\textbf{Dual-level Feature Search Space.} The search space defines what neural architectures might be discovered in principle \cite{liu2018darts, wu2019fbnet}, which plays a crucial role in high-performance NAS. Similarly, NFS is also closely related to a well-designed feature search space that covers as many as possible identity-related cues. As discriminative features are mainly extracted by the shared part of the baseline model \cite{wu2017rgb}, we establish a search space including all feature candidates extracted by every shared convolutional block. More formally, a shared block $L$ takes a feature map $X\in\mathbb{R}^{C_{in} \times W \times H}$ as input and outputs another fine-grained feature map $Y_{L}\in\mathbb{R}^{C_{out} \times \frac{W}{2} \times \frac{H}{2}}$, i.e.,
\begin{equation}
   Y_{L}(p) = \sum_{p^{'}\in R_{k}} W_{c}(p^{'})X(p+p^{'})\quad p\in\Omega,
\end{equation}
where $C$, $W$, and $H$ represent the number of channels, width, and height, respectively. $p$ denotes a specific pixel position, $R_{k}$ is the support region of kernel with size $k$, $W_{c} \in \mathbb{R}^{C_{in}\times C_{out}\times k\times k}$ represents convolution weights, $\Omega=\{(i,j)|i\leq W, j\leq H, i, j \in \mathbb{Z}^{+}\}$ is the spatial domain of $Y_{L}$. The union of $Y_{L}$ forms the vanilla feature space $Y=\{Y_{L}|L\in\{1,...,N\}\}$, and $N$ is the number of blocks. 

Motivated by the fact that discriminative features present modality-specific distributions in spatial and channel dimensions of $Y_{L}$ \cite{choi2020hi}, we introduce modality-aware search cells to decompose the original feature space $Y$ into two subspaces: pixel-level subspace and depth-level subspace. The former includes vectors in each spatial position that describe local patches of an input image. The latter contains multiple detector response maps that globally reflect particular properties of the image content. Fig. \ref{fig:pipeline}(Left) illustrates how the search cell exactly works. Given the output feature map $Y_{L}$, we first initialize a set of parameters $\mathcal{P}^{m}_{L}$ with a uniform distribution to map features from modality $m$ into a specific probability field. As for pixel-level feature search, $\mathcal{P}^{m}_{L}\in\mathbb{R}^{C_{out}\times\frac{W}{2}\times\frac{H}{2}}$ covers every pixel in the spatial domain. And for depth-level search,  $\mathcal{P}^{m}_{L}\in\mathbb{R}^{C_{out}}$ contains all channels of $Y_{L}$. During the searching process, the probability field is activated by a sigmoid function, denoted as $\tilde{\mathcal{P}}^{m}_{L}=\sigma(\mathcal{P}^{m}_{L})$, to indicate the possibilities of features at corresponding positions are informative to distinguish different persons. Then, a binary search gate $\mathcal{G}^{m}_{L}(p)$ is generated based on $\tilde{\mathcal{P}}^{m}_{L}$ to determine whether the pixel at position $p$ should be selected. By passing $Y_{L}$ through all search gates, the output activation map $\tilde{Y}_{L}$ can be formulated as:
\begin{equation}
\tilde{Y}_{L}(p)=
\begin{cases}
Y_{L}(p),& \text{$\mathcal{G}^{m}_{L}(p)=1$}\\
0,& \text{$\mathcal{G}^{m}_{L}(p)=0$}.
\end{cases}
\end{equation}

Here, Eq. \ref{eq:2} is transformed into an optimization problem with all search gates $\mathcal{G}^{m}$ as the upper-level variables and the network weights $W$ as lower-level ones, that is:
\begin{equation}
\label{eq:5}
\begin{array}{ll}
\min \limits_{\mathcal{G}^{m}} & \mathcal{L}_{v a l}\left(W^{*}(\mathcal{G}^{m}), \mathcal{G}^{m}\right) \\
\text { s.t. } & W^{*}(\mathcal{G}^{m})=\arg\underset{W}{\min} \ \mathcal{L}_{train}(W, \mathcal{G}^{m}),
\end{array}
\end{equation}
where $\mathcal{L}_{train}$ and $\mathcal{L}_{val}$ denote the training loss and the validation loss (Eq. \ref{eq:12}), respectively.

\textbf{Search Algorithm.} As the search space is discrete and large-scale, finding the optimal feature set through brute-force enumeration is much inefficient. To tackle this obstacle, we utilize reparameterization tricks \cite{liu2018darts} to relax the search space to be continuous and directly improve the search efficiency via SGD. We assume that feature selection is essentially a binary classification problem and thus exploiting a continuous Bernoulli distribution \cite{loaiza2019continuous} to simulate stochastic discrete sampling with $\tilde{\mathcal{P}}^{m}\in(0, 1)$. Based on $\mathcal{G}^{m}$, the sampled features $\mathcal{X}^{m}$ are as:
\begin{equation}\small
\begin{aligned}
\mathcal{X}^{m}\sim Bernoulli (\tilde{\mathcal{P}}^{m}) \iff& p({x}^{m}\mid\tilde{\mathcal{P}}^{m})\propto\hat{p}({x}^{m}\mid\tilde{\mathcal{P}}^{m})\\
&={(\tilde{\mathcal{P}}^{m}})^{x^{m}}(1-\tilde{\mathcal{P}}^{m})^{1-x^{m}},\\
\end{aligned}
\end{equation}

\begin{equation}\small
p({x}^{m}\mid\tilde{\mathcal{P}}^{m})=C(\tilde{\mathcal{P}}^{m}){(\tilde{\mathcal{P}}^{m}})^{x^{m}}(1-\tilde{\mathcal{P}}^{m})^{1-x^{m}},
\end{equation}
\begin{equation}
C(\tilde{\mathcal{P}}^{m})=
\begin{cases}
\frac{2\tanh^{-1}(1-2\tilde{\mathcal{P}}^{m})}{1-2\tilde{\mathcal{P}}^{m}}, \quad &\text{if $\tilde{\mathcal{P}}^{m}\neq0.5$}\\
2, \quad &\text{otherwise.} 
\end{cases}
\end{equation}
After relaxation, $\mathcal{G}^{m}$ and $W$ can be jointly optimized using the straight-through estimator (STE) \cite{bengio2013estimating}:
\begin{equation}
\nabla_{\mathcal{G}^{m}}\mathcal{L}_{val} \approx \nabla_{\tilde{\mathcal{P}}^{m}}\mathcal{L}_{val}.
\end{equation}

Finally, Eq. \ref{eq:5} is solved with Neural Feature Search outlined in Alg. \ref{arg:NFS}. We first search for identity-related features by iteratively optimizing $\mathcal{G}^{m}$ with $\mathcal{L}_{val}$ and $W$ with $\mathcal{L}_{train}$. After obtaining the optimal search gates,  the whole network is trained and evaluated following the standard feature learning paradigm of RGB-IR ReID \cite{ye2020dynamic}. 

\begin{algorithm}[h]
\caption{NFS - Neural Feature Search}
\label{arg:NFS}
\textbf{Input:} the search parameters $\tilde{\mathcal{P}}^{m}$ and the network weights $W$; the training set $\mathbb{D}_{T}$ and the testing set $\mathbb{D}_{E}$\\
\textbf{Output:} the trained network and the optimal feature set

\begin{algorithmic}[1]\small
\State Randomly split $\mathbb{D}_{T}$ into the search training set $\mathbb{D}_{train}$ and the search validation set $\mathbb{D}_{val}$
\While{\textit{not converged}}
\State Update $W$ by descending $\nabla_{W}\mathcal{L}_{train}(W, \tilde{\mathcal{P}}^{m})$ on $\mathbb{D}_{train}$
\State Update search gates $\mathcal{G}^{m}$ by descending 
\Statex\quad\quad$\nabla_{\tilde{\mathcal{P}}^{m}}\mathcal{L}_{val}(W-\nabla_{W}\mathcal{L}_{train}(W,  \tilde{\mathcal{P}}^{m}), \tilde{\mathcal{P}}^{m})$) on $\mathbb{D}_{val}$
\EndWhile
\State Derive optimal $\mathcal{G}^{m}$ based on $\tilde{\mathcal{P}}^{m}$
\State Train the network weight $W$ with the derived $\mathcal{G}^{m}$ on $\mathbb{D}_{T}$
\State Evaluate the network and $\mathcal{G}^{m}$ on $\mathbb{D}_{E}$
\end{algorithmic}
\end{algorithm}

\vspace{-0.3cm}
\subsection{Cross-Modality Contrastive Optimization}
\label{sec:3.3}
Apart from the search efficiency, how to supervise search cells to select more informative features is also important for NFS. Unlike close-set classification tasks, RGB-IR ReID is an open-set problem where identities in testing are different from those in training. In such a scenario, the selected features of `seen' and `unseen' classes may be tangled in the feature space.  Meanwhile, the appearance discrepancy between RGB and IR images often enlarges the feature distribution variance of each class, leading to fuzzy decision boundaries in identity recognition problems.

Here, we attend to decrease the feature distribution variance from an invariant feature selection perspective. To this end, we introduce a ReID-oriented optimization criterion that can eliminate modality discrepancy and maximize the inter-class distance simultaneously. The basic idea comes from recent advances on contrastive learning \cite{tian2019contrastive, khosla2020supervised, he2020momentum}, which aim to attract positive pairs whilst repulsing negative ones \cite{chen2020simple}. Given a training batch $B=\{(i_{rgb}, i_{ir}) | i_{rgb} \in \mathcal{I}_{rgb}, i_{ir} \in \mathcal{I}_{ir}\}$, the half of which are RGB images $\mathcal{I}_{rgb}$ while the others are their IR counterparts $\mathcal{I}_{ir}$, we randomly arrange their embedding vectors $\vec{i}_{rgb}$ and $\vec{i}_{ir}$ into multiple cross-modality pairs $(\vec{i}_{rgb}, \vec{i}_{ir})$ and generate pair-wise labels according to their identities $ID(\vec{i}_{rgb})$ and $ID(\vec{i}_{rgb})$:
\begin{equation}
Label = 
\begin{cases}
1,& \text{$ID(\vec{i}_{rgb}) = ID(\vec{i}_{ir})$}\\
0,& \text{$ID(\vec{i}_{rgb}) \neq ID(\vec{i}_{ir})$}.
\end{cases}
\end{equation}

For each positive image pair, we seek to minimize the distance between them, so that the modality discrepancy and intra-class variations can be jointly eliminated. We evaluate the pair-wise distance in Euclidean space, which is widely applied in image retrieval \cite{hadsell2006dimensionality, ye2020deep}, i.e.,
\begin{equation}
D_{E}=||\vec{i}_{rgb} - \vec{i}_{ir}||_{2}.
\end{equation}
On the contrary, for negative pairs, we aim to keep them far from each other for distinction. In order to make the optimization objective explicitly, we quantify the dissimilarity of each negative pair $D_{T}$ with an explicit margin $T$:
\begin{equation}
\label{eq:10}
D_{T}=\max(0, T-||\vec{i}_{rgb} - \vec{i}_{ir}||_{2}).
\end{equation}
Taking all positive and negative pairs into account, the contrastive loss $\mathcal{L}_{c}$ can be formulated as:
\begin{equation}
\label{eq:11}
\mathcal{L}_{c}(Label, \vec{i}_{rgb}, \vec{i}_{ir}) =(Label)(D_{E})^{2} + (1-Label)(D_{T})^{2}.
\end{equation}

The overall learning objective for NFS is a weighted summation of the baseline loss $\mathcal{L}_{b}$ and cross-modality contrastive loss $\mathcal{L}_{c}$, defined as:
\begin{equation}
\label{eq:12}
\mathcal{L} = \mathcal{L}_{b} + \lambda \mathcal{L}_{c},
\end{equation}
where $\lambda$ is a trade-off coefficient to balance the influence of each learning objective.

\section{Experiments}
\begin{table*}[t]\scriptsize
\caption{Comparison on the SYSU-MM01 dataset with Rank-1, 10, 20 (\%) and mAP (\%) evaluation metrics.}
\label{Table:SYSU}
\begin{tabular}{ccccccccccccccccc}
\hline
\multicolumn{1}{c|}{\multirow{3}{*}{Method}} & \multicolumn{8}{c|}{All Search}                                   & \multicolumn{8}{c}{Indoor Search}                                \\ \cline{2-17} 
                        & \multicolumn{4}{|c}{Single-shot} & \multicolumn{4}{c|}{Multi-shot} & \multicolumn{4}{c}{Single-shot} & \multicolumn{4}{c}{Multi-shot} \\ \cline{2-17} 
\multicolumn{1}{c|}{}                        & r1     & r10    & r20   & \multicolumn{1}{l|}{mAP}   & r1     & r10   & r20   & \multicolumn{1}{l|}{mAP}   & r1     & r10    & r20   & \multicolumn{1}{l|}{mAP}   & r1     & r10   & r20   & mAP   \\ \hline
\multicolumn{1}{c|}{HOG \cite{dalal2005histograms}}                     & 2.76   & 18.3   & 31.9  & \multicolumn{1}{l|}{4.24}  & 3.82   & 22.8  & 37.6  & \multicolumn{1}{l|}{2.16}  & 3.22   & 24.7   & 44.5  & \multicolumn{1}{l|}{7.25}  & 4.75   & 29.2  & 49.4  & 3.51  \\
\multicolumn{1}{c|}{LOMO \cite{liao2015person}}                    & 3.64   & 23.2   & 37.3  & \multicolumn{1}{l|}{4.53}  & 4.70   & 28.2  & 43.1  & \multicolumn{1}{l|}{2.28}  & 5.75   & 34.4   & 54.9  & \multicolumn{1}{l|}{10.2}  & 7.36   & 40.4  & 60.3  & 5.64  \\
\multicolumn{1}{c|}{Zero-Padding \cite{wu2017rgb}}            & 14.8   & 54.1   & 71.3  & \multicolumn{1}{l|}{15.9}  & 19.1   & 61.4  & 78.4  & \multicolumn{1}{l|}{10.9}  & 20.6   & 68.4   & 85.8  & \multicolumn{1}{l|}{26.9}  & 24.4   & 75.9  & 91.3  & 18.6  \\
\multicolumn{1}{c|}{TONE+HCML \cite{ye2018hierarchical}}               & 14.3   & 53.2   & 69.2  & \multicolumn{1}{l|}{16.2}  & -      & -     & -     & \multicolumn{1}{l|}{-}     & -      & -      & -     & \multicolumn{1}{l|}{-}     & -      & -     & -     & -     \\
\multicolumn{1}{c|}{BDTR \cite{ye2018visible}}                    & 17.0   & 55.4   & 72.0  & \multicolumn{1}{l|}{19.7}  & -      & -     & -     & \multicolumn{1}{l|}{-}     & -      & -      & -     & \multicolumn{1}{l|}{-}     & -      & -     & -     & -     \\
\multicolumn{1}{c|}{D-HSME \cite{hao2019hsme}}                  & 20.7   & 62.8   & 78.0  & \multicolumn{1}{l|}{23.2}  & -      & -     & -     & \multicolumn{1}{l|}{-}     & -      & -      & -     & \multicolumn{1}{l|}{-}     & -      & -     & -     & -     \\
\multicolumn{1}{c|}{IPVT+MSR \cite{kang2019person}}                & 23.2   & 51.2   & 61.7  & \multicolumn{1}{l|}{22.5}  & -      & -     & -     & \multicolumn{1}{l|}{-}     & -      & -      & -     & \multicolumn{1}{l|}{-}     & -      & -     & -     & -     \\
\multicolumn{1}{c|}{cmGAN \cite{dai2018cross}}                   & 27.0   & 67.5   & 80.6  & \multicolumn{1}{l|}{27.8}  & 31.5   & 72.7  & 85.0  & \multicolumn{1}{l|}{22.3}  & 31.6   & 77.2   & 89.2  & \multicolumn{1}{l|}{42.2}  & 37.0   & 80.9  & 92.1  & 32.8  \\
\multicolumn{1}{c|}{D$^{2}$RL \cite{wang2019learning}}                     & 28.9   & 70.6   & 82.4  & \multicolumn{1}{l|}{29.2}  & -      & -     & -     & \multicolumn{1}{l|}{-}     & -      & -      & -     & \multicolumn{1}{l|}{-}     & -      & -     & -     & -     \\
\multicolumn{1}{c|}{DGD+MSR \cite{feng2019learning}}                 & 37.4   & 83.4   & 93.3  & \multicolumn{1}{l|}{38.1}  & 43.9   & 86.9  & 95.7  & \multicolumn{1}{l|}{30.5}  & 39.6   & 89.3   & 97.7  & \multicolumn{1}{l|}{50.9}  & 46.6   & 93.6  & 98.8  & 40.1  \\
\multicolumn{1}{c|}{JSIA-ReID \cite{wang2020cross}}               & 38.1   & 80.7   & 89.9  & \multicolumn{1}{l|}{36.9}  & 45.1   & 85.7  & 93.8  & \multicolumn{1}{l|}{29.5}  & 43.8   & 86.2   & 94.2  & \multicolumn{1}{l|}{52.9}  & 52.7   & 91.1  & 96.4  & 42.7  \\
\multicolumn{1}{c|}{AlignGAN \cite{wang2019rgb}}                & 42.4   & 85.0   & 93.7  & \multicolumn{1}{l|}{40.7}  & 51.5   & 89.4  & 95.7  & \multicolumn{1}{l|}{33.9}  & 45.9   & 87.6   & 94.4  & \multicolumn{1}{l|}{54.3}  & 57.1   & 92.7  & 97.4  & 45.3  \\
\multicolumn{1}{c|}{AGW \cite{ye2020deep}}                & 47.50   & 84.39   & 92.14  & \multicolumn{1}{l|}{47.65}  & -   & -  & -  & \multicolumn{1}{l|}{-}  & 54.17   & 91.14   & 95.98  & \multicolumn{1}{l|}{62.97}  & -   & -  & -  & -  \\
\multicolumn{1}{c|}{Xmodal \cite{li2020infrared}}                & 49.92   & 89.79   & 95.96  & \multicolumn{1}{l|}{50.73}  & -   & -  & -  & \multicolumn{1}{l|}{-}  & -   & -   & -  & \multicolumn{1}{l|}{-}  & -   & -  & -  & -  \\
\multicolumn{1}{c|}{DDAG \cite{ye2020dynamic}}                & 54.75   & 90.39   & 95.81  & \multicolumn{1}{l|}{53.02}  & -   & -  & -  & \multicolumn{1}{l|}{-}  & 61.02   & 94.06   & 98.41  & \multicolumn{1}{l|}{67.98}  & -   & -  & -  & -  \\
\hline
\multicolumn{1}{c|}{NFS (Ours) }                & \textbf{56.91}   & \textbf{91.34}   & \textbf{96.52}  & \multicolumn{1}{l|}{\textbf{55.45}}  & \textbf{63.51}   & \textbf{94.42}  & \textbf{97.81}  & \multicolumn{1}{l|}{\textbf{48.56}}  & \textbf{62.79}   & \textbf{96.53}   & \textbf{99.07}  & \multicolumn{1}{l|}{\textbf{69.79}}  & \textbf{70.03}   & \textbf{97.70}  & \textbf{99.51}  & \textbf{61.45}  \\
\hline
\end{tabular}
\end{table*}
\subsection{Datasets and Experimental Settings}
\textbf{Datasets.} Our experiments are based on two standard real-world benchmarks for RGB-IR person ReID, named SYSU-MM01 \cite{wu2017rgb} and RegDB \cite{nguyen2017person}, respectively. The SYSU-MM01 dataset contains images captured by four visible and two near infrared cameras in indoor and outdoor environments. Statistically, the training set includes 22,258 RGB and 11,909 IR images of 395 identities, and the query set involves 3,803 IR images of 96 identities. The gallery set has four versions according to different evaluation modes, including \textit{all search} or \textit{indoor search} and \textit{single-shot} or \textit{multi-shot}. Details of each mode can be found in \cite{wu2017rgb}. The RegDB dataset contains 8,240 images of 412 identities, with 206 identities for training and the rest for testing. Each identity has 10 IR and 10 RGB images. We evaluate both \textit{visible-to-infrared} and \textit{infrared-to-visible} modes \cite{wang2019rgb} by alternatively using all RGB/IR images as the gallery set.

\begin{table*}[]\scriptsize
\caption{Comparison on the RegDB dataset with Rank-1, 10, 20 (\%) and mAP (\%) evaluation metrics.}
\label{Table:RegDB}
\begin{center}
\begin{tabular}{c|cccc|cccc}
\hline
\multirow{2}{*}{Method} & \multicolumn{4}{c|}{\textit{Visible to Infrared}} & \multicolumn{4}{c}{\textit{Infrared to Visible}} \\ \cline{2-9} 
                        & r1         & r10        & r20        & mAP        & r1         & r10        & r20        & mAP       \\ \hline
Zero-Padding \cite{wu2017rgb}            & 17.75      & 34.21      & 44.35      & 18.90      & 16.63      & 34.68      & 44.25      & 17.82     \\
Tone + HCML \cite{ye2018hierarchical}             & 24.44      & 47.53      & 56.78      & 20.88      & 21.70      & 45.02      & 55.58      & 22.24     \\
BDTR \cite{ye2018visible}                    & 33.56      & 58.61      & 67.43      & 32.76      & 32.92      & 58.46      & 68.43      & 31.96     \\
D$^{2}$RL \cite{wang2019learning}                   & 43.4       & 66.1       & 76.3       & 44.1       & -          & -          & -          & -         \\
DGD+MSR \cite{feng2019learning}                & 48.43      & 70.32      & 79.95      & 48.67      & -          & -          & -          & -         \\
JSIA-ReID \cite{wang2020cross}               & 48.1       & -          & -          & 48.9       & 48.5       & -          & -          & 49.3      \\
D-HSME \cite{hao2019hsme}                  & 50.85      & 73.36      & 81.66      & 47.00      & 50.15      & 72.40      & 81.07      & 46.16     \\
IPVT+MSR \cite{kang2019person}                & 58.76      & 85.75      & 90.27      & 47.85      & -          & -          & -          & -         \\
AlignGAN \cite{wang2019rgb}                & 57.9       & -          & -          & 53.6       & 56.3       & -          & -          & 53.4      \\
Xmodal \cite{li2020infrared}               & 62.21      & 83.13      & 91.72      & 60.18      & -          & -          & -          & -         \\
DDAG \cite{ye2020dynamic}                    & 69.34      & 86.19      & 91.49      & 63.46      & 68.06      & 85.15      & 90.31      & 61.80     \\
\hline
NFS (Ours)                    &\textbf{80.54}            &\textbf{91.96}            &\textbf{95.07}            &\textbf{72.10}            &\textbf{77.95}            &\textbf{90.45}            &\textbf{93.62}            &\textbf{69.79}          \\ \hline
\end{tabular}
\end{center}
\end{table*}

\textbf{Evaluation Protocols.} We follow standard evaluation protocols \cite{wu2017rgb, ye2020deep} for RGB-IR ReID. Gallery and query images are from different modalities. The standard cumulated matching characteristics (CMC) curve and mean average precision (mAP) are used for performance evaluation.

\textbf{Implementation Details.} The proposed method is implemented in PyTorch and trained on an NVIDIA 2080Ti GPU. In order to facilitate comparisons with state-of-the-art ReID researches \cite{ye2020deep, ye2020dynamic, lu2020cross}, we adopt the ResNet-50 \cite{he2016deep} pretrained on ImageNet \cite{deng2009imagenet} as our backbone network. Following \cite{ye2020dynamic, ye2019bi, ye2020cross, ye2020deep}, we set the stride of the last convolutional block as 1 for fine-grained feature maps. All images are resized to 288 $\times$ 144 then augmented with random cropping and horizontal flipping. We randomly sample 80\% images from the original training set as the search training set and use the rest as the search validation set (Alg. \ref{arg:NFS}, Line 1). We first make all search cells learnable to discover the optimal discriminative feature set. After obtaining the optimal feature set, we retrain the network on the original training set. At the training stage, we adopt a warm-up strategy \cite{luo2019strong} and optimize the two-stream CNN using SGD with 0.9 momentum during a total of 80 epochs. The initial learning rate is set to 0.1 and decays by 0.1 and 0.01 at the 16th and 50th epoch, respectively. Following \cite{ye2020dynamic}, we randomly sample 8 identities with 4 RGB and 4 IR images per person, resulting in totally 64 images for each training batch.

\subsection{Comparison with State-of-the-art Methods}
In this subsection, we compare the proposed NFS with naive baselines as well as the state-of-the-art methods, including traditional feature extraction methods (HOG \cite{dalal2005histograms} and LOMO \cite{liao2015person}); GAN-based models (cmGAN \cite{dai2018cross}, D$^{2}$RL \cite{wang2019learning}, JSIA-ReID \cite{wang2020cross}, AlignGAN \cite{wang2019rgb}, and Xmodal \cite{li2020infrared}); deep metric learning (BDTR \cite{ye2018visible}, D-HSME \cite{hao2019hsme}, IPVT+MSR \cite{kang2019person}, and DGD+MSR \cite{feng2019learning}); and shared feature learning approaches (Zero-Padding \cite{wu2017rgb}, TONE+HCML \cite{ye2018hierarchical}, AGW \cite{ye2020deep}, and DDAG \cite{ye2020dynamic}). Since most of them follow the standard evaluation protocols of the two experimental datasets, we directly use the original results from published papers for comparison.

Experimental results on SYSU-MM01 are shown in Table \ref{Table:SYSU}. We see that there is a significant performance decline when applying hand-crafted descriptors HOG and LOMO to cross-modality ReID, regardless of their promising capacities in general ReID tasks. Besides, image synthesis methods (AlignGAN, JSIA-ReID, Xmodal, and D$^{2}$RL) perform better than traditional shared feature learning approaches (Zero-Padding and TONE+HCML), possibly owing to the effectiveness of pixel-level alignment. Specifically, recent methods such as AGW, DDAG, as well as our proposed NFS outperform typical GAN-based approaches. This is probably because it is ill-posed to transfer identity-related information of IR images to generated RGB images. Notably, the proposed model achieves 56.91\% Rank-1 identification rate and 55.45\% mAP score in the most difficult \textit{single-shot \& all search} setting, which outperforms most of state-of-the-art methods by a large margin. Compared to DDAG based on the graph attention mechanism, NFS is much easier to implement and still presents better performance. Similar improvement can be observed in \textit{multi-shot} modes. For example, our method largely surpasses AlignGAN with the improvement of 12.01\% in Rank-1 and 14.66\% in mAP, which demonstrates highly robustness when the gallery size increases. 

Results on the RegDB dataset are listed in Table \ref{Table:RegDB}. Generally, performance of all methods is higher than that on SYSU-MM01, as images of RegDB present less intra-class variations \cite{wang2019rgb}. Our approach greatly improves the state-of-the-art under both evaluation modes. Specifically, in the \textit{visible-to-infrared} mode, NFS makes a marked improvement of 11.20\% in Rank-1 and 8.64\% in mAP compared to the top-performing method DDAG \cite{ye2020dynamic}. Similar increment also presents in the \textit{infrared-to-visible} mode, which shows that our method is robust to multi-modal query settings. In conclusion, the above results clearly indicate the effectiveness of our automated feature search paradigm.

\subsection{Ablation Study}
This subsection studies the effectiveness of each module involved in NFS on SYSU-MM01 ( \textit{all} and \textit{indoor search}, \textit{single-shot} settings). As in Table \ref{Table:ablation}, $\mathcal{B}$ denotes the baseline two-stream network with the learning objective $\mathcal{L}_{b}$, $\mathcal{N}$ represents the neural feature search block, and $\mathcal{C}$ indicates the cross-modality contrastive optimization mechanism.
\begin{table}[h]\scriptsize
\caption{Evaluation of each module on SYSU-MM01.}
\label{Table:ablation}
\begin{center}
\begin{tabular}{c|ccc|ccc}
\hline
\multirow{2}{*}{Method} & \multicolumn{3}{c|}{All Search} & \multicolumn{3}{c}{Indoor Search} \\ \cline{2-7} 
                        & r1        & r10      & mAP      & r1        & r10       & mAP       \\ \hline
$\mathcal{B}$                       & 47.00     & 84.11    & 46.46    & 52.70     & 89.30     & 60.93     \\
$\mathcal{B}+\mathcal{N}$                     & 48.91          & 86.03         & 47.92         & 54.12          & 92.20          & 62.31          \\
$\mathcal{B}+\mathcal{C}$                     & 52.29          & 90.22         & 50.91         & 57.23           & 94.14          & 65.32          \\
$\mathcal{B}+\mathcal{N}+\mathcal{C}$                  & \textbf{56.91}           & \textbf{91.34}         & \textbf{55.45}         & \textbf{62.79}           & \textbf{96.53}          & \textbf{69.79}          \\ \hline
\end{tabular}
\end{center}
\end{table}
\vspace{-0.5cm}

\textbf{Effectiveness of Neural Feature Search.} We evaluate how much improvement can be made by NFS with baseline learning objective $\mathcal{L}_{b}$. To be fair, all hyperparameters are fixed during evaluation. As shown in $2^{nd}$ row of Table \ref{Table:ablation}, NFS brings 1.91\% Rank-1 and 1.46\% mAP increases in \textit{all search} mode compared with $\mathcal{B}$ (row 2). Similar improvement can be observed in \textit{indoor search} mode. This increment suggests that automated feature selection not only governs the baseline network to focus on informative parts of human bodies but also filters high sample noises.

\textbf{Influence of Contrastive Optimization.} Here, we investigate the contribution of contrastive loss. Considerable enhancement (5.29\% of Rank-1 and 4.45\% of mAP for \textit{all search}, 4.53\% of Rank-1 and 4.39\% of mAP for \textit{indoor search}) on the baseline model can be observed in Table \ref{Table:ablation}. This improvement manifests the superiority of contrastive loss for learning identity-related information. We further validate its effectiveness on NFS and the results are listed in the $4^{th}$ row of Table \ref{Table:ablation}. We observe that, with contrastive loss, NFS significantly surpasses the baseline model with 9.91\% growth of Rank-1 and 8.99\% gain of mAP for \textit{all search}, while performance boost in \textit{indoor search} is even more pronounced. Notably, from the comparison between $3^{rd}$ and $4^{th}$ row of Table \ref{Table:ablation}, we see that NFS brings more benefits (4.62\% of Rank-1 and 4.54\% of mAP) to the baseline model with $\mathcal{C}$, demonstrating that the contrastive loss not only contributes to the optimization of $\mathcal{B}$, but also encourages NFS to discover more discriminative features.

\textbf{Impact of Search Scope.} In this part, we compare the performance of NFS conducted at different convolution stages (Table \ref{Table:stage}). The Rank-1 and mAP tend to increase when NFS is conducted on more layers. The best result appears when we perform NFS on Stage 1, 2, and 3. This is probably because, searching more stages expands the search space, which allows us to explore more varied selections of features. However, blindly extending the search scope will pose great challenges to the discovery of an optimal feature set. The underlying reason is that STE will generate more and more gradient estimation errors when backpropagating through too many layers \cite{bengio2013estimating}.
\begin{table}[h]\scriptsize
\caption{Comparison on NFS at different convolution stages.}
\label{Table:stage}
\begin{center}
\begin{tabular}{c|cccc|cc}
\hline
   & Stage1 & Stage2 & Stage3 & Stage4 & r1 & mAP \\ \hline
1  & $\surd$      &   -     &   -     &   -     &53.75    &53.02     \\
2  &  -      & $\surd$      &  -      &  -      & 53.41    & 52.67    \\
3  & -       &    -    & $\surd$      &  -      & 53.64   & 53.25    \\
4  & -       &    -    &    -    & $\surd$      & 53.92   & 52.83    \\ \hline
5  & $\surd$      & $\surd$      &  -      &   -     &55.62    & 54.31    \\
6  & $\surd$      &  -      & $\surd$      &   -     & 54.85   & 53.97    \\
7  & $\surd$      &  -      &   -     & $\surd$      & 54.45   &  53.76   \\
8  &   -     & $\surd$      & $\surd$      &  -      & 55.13   &  54.07   \\
9  &   -     & $\surd$      &  -      & $\surd$      & 54.41   &  53.29   \\
10 &   -     &   -     & $\surd$      & $\surd$      & 54.42   &  53.87   \\ \hline
11 & $\surd$      & $\surd$      & $\surd$      &   -     & \textbf{56.91}   & \textbf{55.45}  \\
12 & $\surd$      & $\surd$      &   -     & $\surd$      & 55.21   &  53.13   \\
13 & $\surd$      &  -      & $\surd$      & $\surd$      & 55.72   & 54.35    \\
14 &   -     & $\surd$      & $\surd$      & $\surd$      & 55.94   & 54.62    \\ \hline
15 & $\surd$      & $\surd$      & $\surd$      & $\surd$      & 55.18   & 54.21    \\ \hline
\end{tabular}
\end{center}
\end{table}
\vspace{-0.3cm}
\subsection{Influence of Hyperparameters} 
In this subsection, we investigate the influence of hyperparameters involved in NFS, including the contrastive margin $T$ (Eq. \ref{eq:10}) and trade-off coefficient $\lambda$ (Eq. \ref{eq:12}). All results are based on SYSU-MM01 (\textit{single-shot \& all search}).

\begin{figure}[h]
\begin{center}
   \includegraphics[width=1\linewidth]{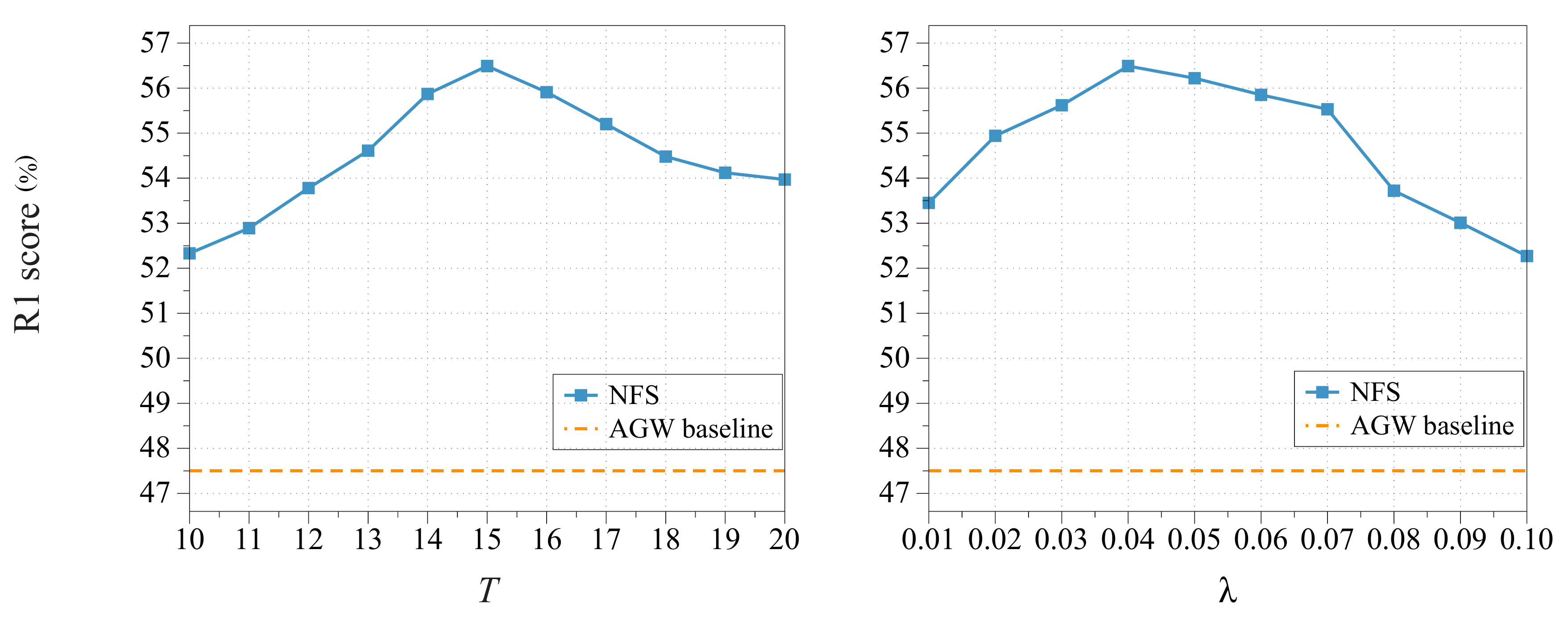}
\end{center}
\vspace{-0.5cm}
   \caption{Parameter analysis for margin $T$ and trade-off weight $\lambda$.}
   \label{fig:hyper}
\end{figure}
\textbf{The Contrastive Margin $T$.} Due to significant appearance differences between RGB and IR images, the original distance among negative pairs is relatively large. Thus, we tune the contrastive margin $T$ from 10 to 20. The corresponding Rank-1 results are shown in Fig. \ref{fig:hyper}(Left). NFS consistently outperforms the AGW baseline \cite{ye2020deep} with different margins and achieves best performance at $T=15$.

\textbf{The Trade-off Coefficient $\lambda$.} We also evaluate influence of the trade-off coefficient $\lambda$. Since the initial contrastive loss value may be very large, we consider $\lambda$ from 0.01 to 0.1. As in Fig. \ref{fig:hyper}(Right), consistent improvement can be observed again when we apply different $\lambda$. NFS achieves the best Rank-1 accuracy when $\lambda=0.04$.
\begin{figure}[h]
\begin{center}
   \includegraphics[width=0.8\linewidth]{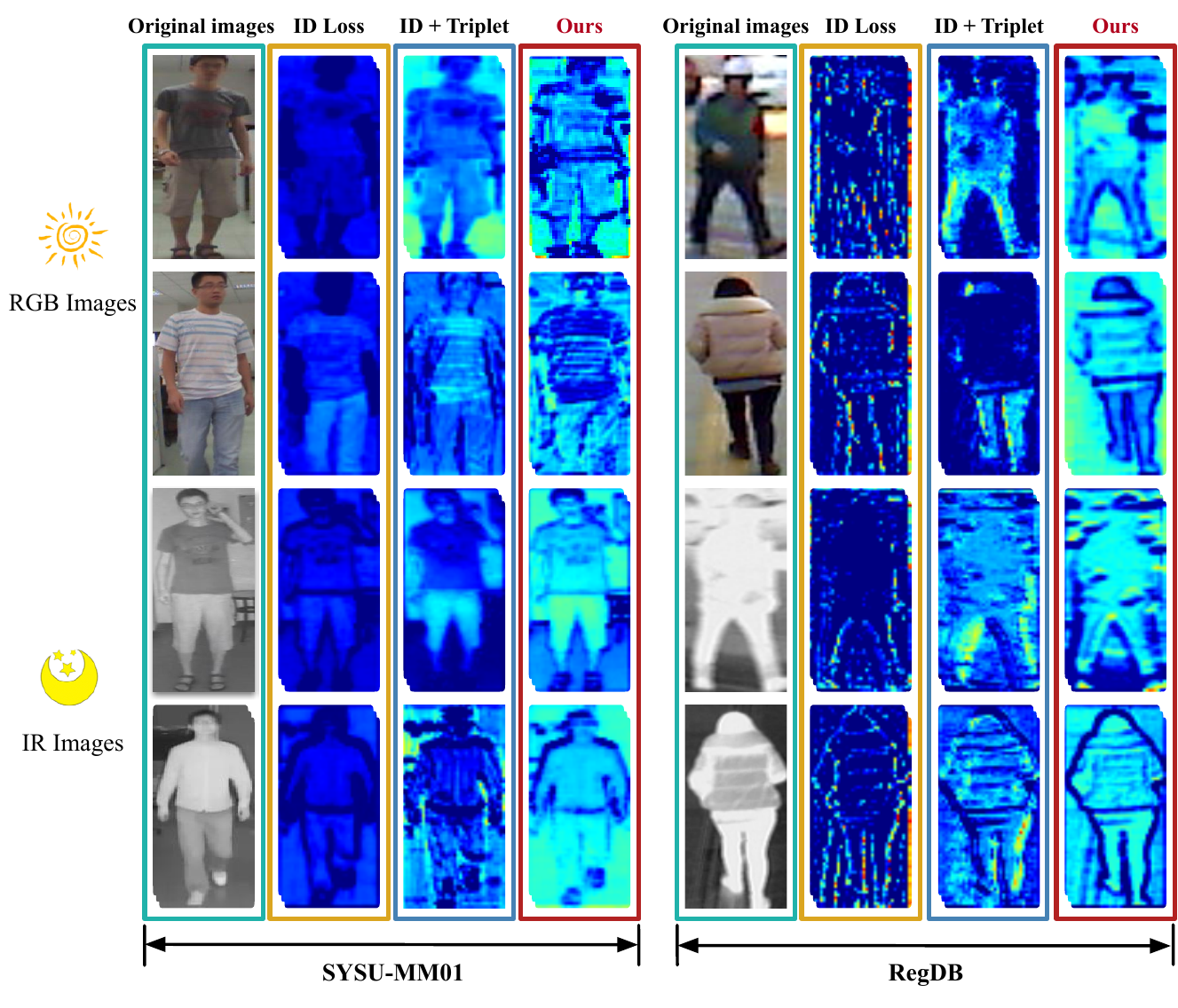}
\end{center}
\vspace{-0.5cm}
   \caption{Visualization of feature maps produced by NFS with different loss functions on SYSU-MM01 and RegDB datasets. Best viewed in color.}
\label{fig:Visualization}
\end{figure}
\subsection{Visualization of Learned Features}
\label{sec:visualization}
In order to inspect the effectiveness of our feature search based method, we visualize feature maps in the first shared block for 8 randomly selected images (4 samples per modality) on the two benchmark datasets (Fig. \ref{fig:Visualization}). It can be observed that, with the introduction of triplet loss (column 3), background noises are effectively eliminated while person information is preserved by search cells. We also see that significant improvement can be achieved when applying the contrastive loss (column 4) to NFS -- not only irrelevant information is further filtered but also more discriminative cues are detected simultaneously.

\begin{figure}[t]
\begin{center}
   \includegraphics[width=1\linewidth]{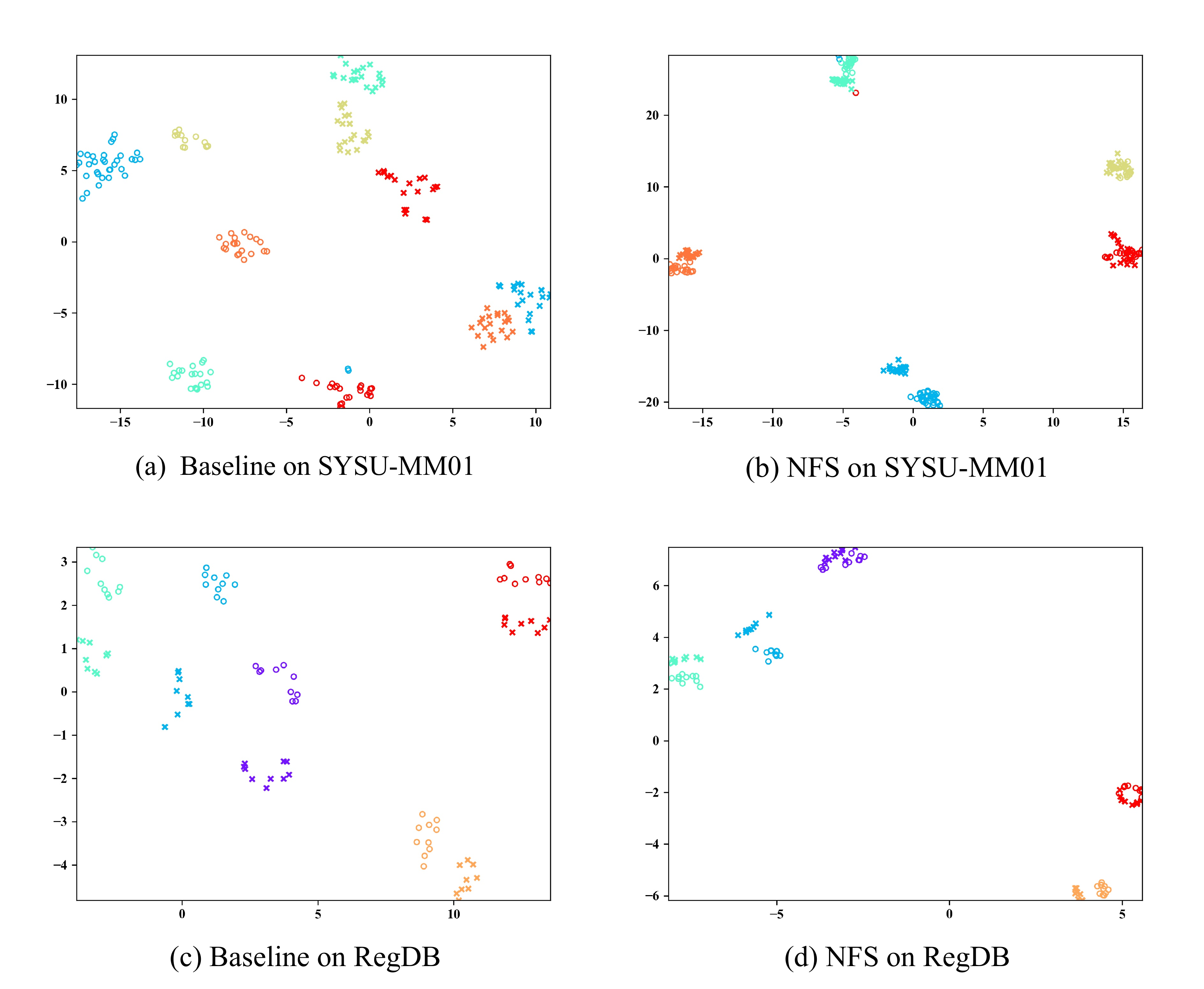}
\end{center}
\vspace{-0.5cm}
\caption{t-SNE visualization of the distribution of learned representations from NFS and the baseline method \cite{ye2020deep}. Different colors represent features of different identities. Circle and cross symbols refer to features of RGB and infrared images, respectively.}
\label{fig:t-SNE}
\end{figure}

We also examine the internal features captured by NFS using t-SNE \cite{maaten2008visualizing}. As shown in Fig. \ref{fig:t-SNE}, we visualize the learned representations of NFS and the baseline method on SYSU-MM01 and RegDB (5 randomly selected person identities per dataset). Specifically, Fig. \ref{fig:t-SNE}(a) and \ref{fig:t-SNE}(c) show the distribution of features extracted by the baseline method while Fig. \ref{fig:t-SNE}(b) and \ref{fig:t-SNE}(d) illustrate the NFS feature distribution. In comparison with Fig. \ref{fig:t-SNE}(a) and \ref{fig:t-SNE}(c), we see that feature distributions from visible and infrared modalities are fairly closer and less discriminable in Fig. \ref{fig:t-SNE}(b) and \ref{fig:t-SNE}(d). This indicates that NFS effectively minimizes the modality gap by aligning distributions of the two modalities. Furthermore, it is also observed that the proposed method separates feature points into disjoint clusters with larger inter-class margin while ensuring positive pairs from different modalities well aggregated. In a nutshell, NFS has a strong capability of detecting more discriminative cues in cross-modality settings.

\section{Conclusion}
This paper presents a novel insight of automated feature selection for RGB-IR ReID. A Neural Feature Search (NFS) paradigm is proposed to adaptively discover more identity characteristics. We first construct a dual-level feature search space, which makes it possible to jointly perform global-channel and local-spatial search operations. Then, we develop an efficient search algorithm to accelerate the selection process. Governed by a cross-modality contrastive optimization objective, this auto-searching algorithm is better able to select more high-quality invariant feature subsets for matching and retrieval. Experimental results on two standard RGB-IR ReID benchmarks demonstrate the effectiveness of NFS surpassing previous state-of-the-arts.
\vspace{0.3cm}

\noindent\textbf{Acknowledgements.} We are grateful to the AC panel and anonymous reviewers for their fruitful comments, corrections, and inspirations. We also thank Chuang Zhang, Yang Du, and Mengyao Tao for helpful discussions.

\newpage

{\small
\bibliographystyle{ieee_fullname}
\bibliography{egbib}
}

\newpage
\section*{The Architecture of Our Baseline Network}
\begin{figure}[h]
\label{fig:two-stream}
\begin{center}
   \includegraphics[width=1\linewidth]{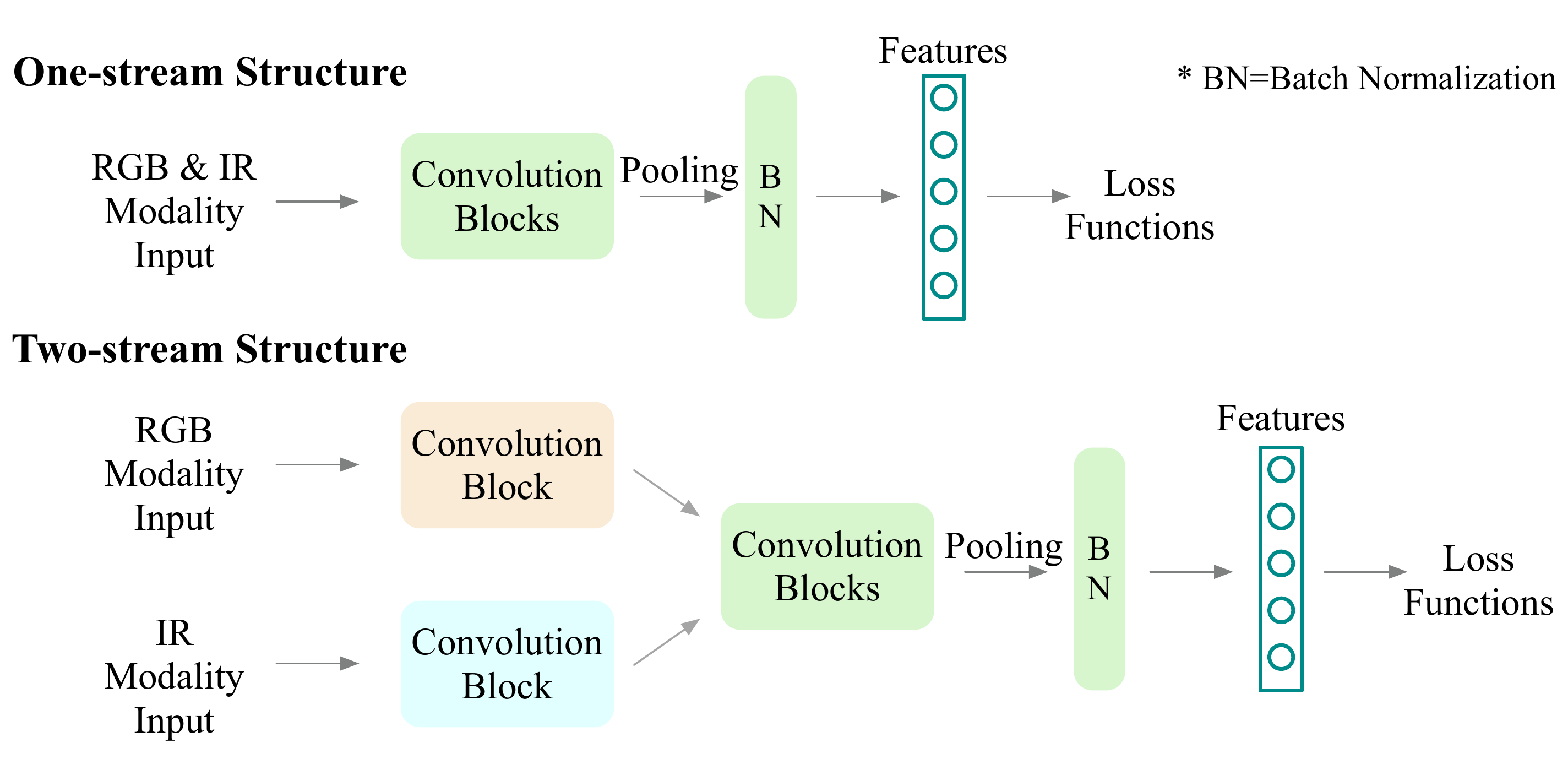}
\end{center}
   \caption{Two-stream structure of the baseline network.}
\end{figure}

In our implementation, the baseline network adopts a commonly used two-stream structure (Fig. \ref{fig:two-stream}) and takes ResNet-50 as the backbone.

Architecture details of the baseline network are shown in Table \ref{table:structure}. All images are resized to 288 $\times$ 144 as the network inputs. The stride of the last convolutional block is set to 1 so as to obtain fine-grained feature maps. The other hyper-parameters are following \cite{he2016deep} without tuning.
\begin{table}[h]\scriptsize
\caption{Architecture details of our two-stream baseline network.}
\vspace{-0.25cm}
\label{table:structure}
\begin{center}
\begin{tabular}{c|c|c|c}
\hline
Layer name             & Output size            & 50-layer & Type \\ \hline
conv1                  & 144$\times$72                 & 7$\times$7, 64, stride 2               &      modality-specific               \\ \hline
\multirow{2}{*}{conv2} & \multirow{2}{*}{72$\times$36} & 3$\times$3 max pool, stride 2          & \multirow{2}{*}{shared}   \\ \cline{3-3}
                       &                        &  \tabincell{c}{\\ $\begin{bmatrix} 1\times1, 64 \\  3\times3, 64 \\ 1\times1, 256 \end{bmatrix} \times 3$ \\ \quad}                               &                     \\ \hline 
conv3                  & 36$\times$18                  &  \tabincell{c}{\\ $\begin{bmatrix} 1\times1, 128 \\  3\times3, 128 \\ 1\times1, 512 \end{bmatrix} \times 4$  \\ \quad}                              &     shared                \\ \hline
conv4                  & 18$\times$9                   &   \tabincell{c}{\\$\begin{bmatrix} 1\times1, 256 \\  3\times3, 256 \\ 1\times1, 1024 \end{bmatrix} \times 6$ \\ \quad}                               &     shared                \\ \hline
conv5                  & 18$\times$9                   &   \tabincell{c}{\\$\begin{bmatrix} 1\times1, 512 \\  3\times3, 512 \\ 1\times1, 2048 \end{bmatrix} \times 3$ \\ \quad}                             &     shared                \\ \hline
\end{tabular}
\end{center}
\end{table}

\section*{Visualization of Retrieved Examples}
\begin{figure*}[t]
\begin{center}
   \includegraphics[width=1\linewidth]{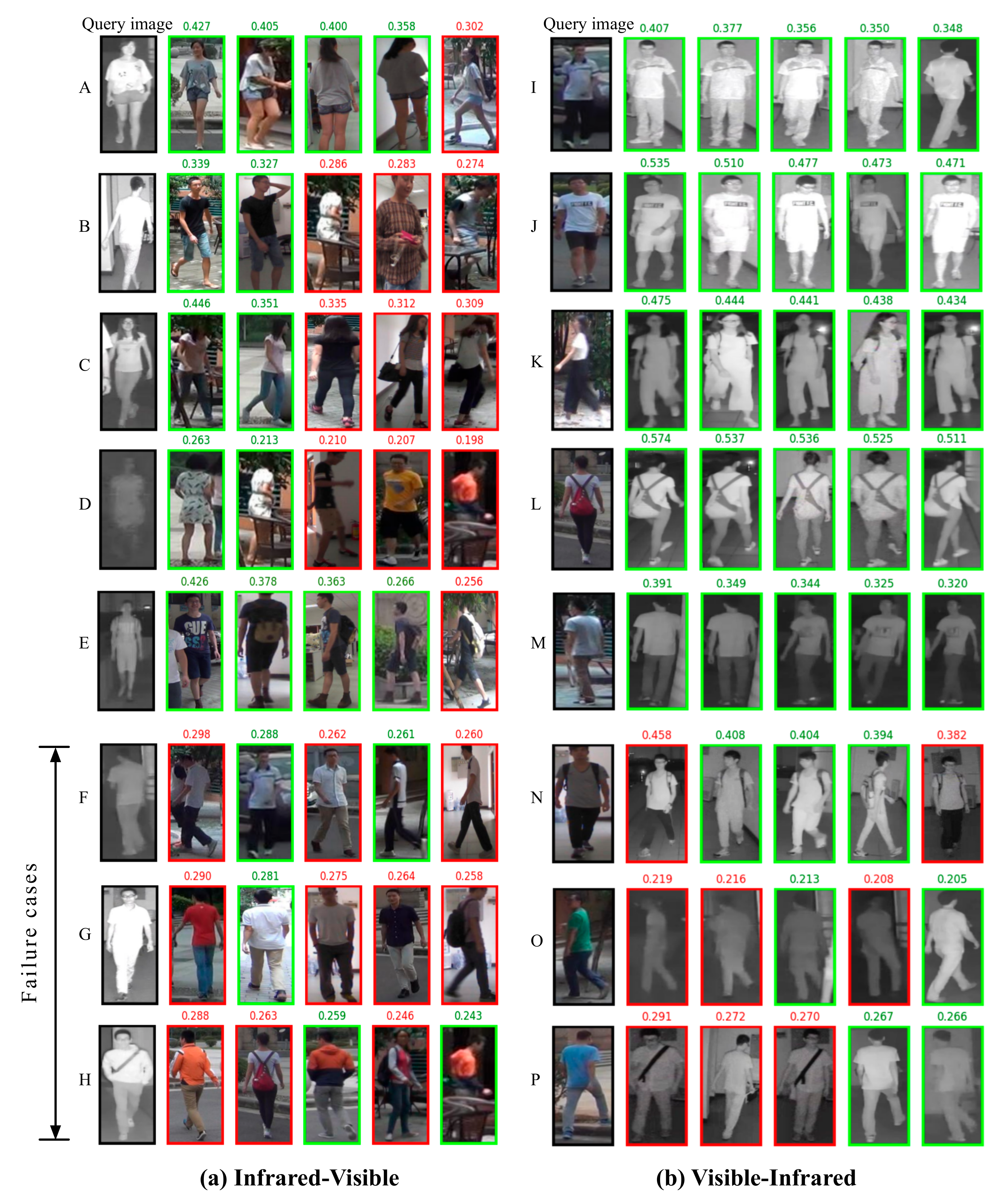}
\end{center}
\vspace{-0.3cm}
\caption{The top-5 retrieval results for 16 randomly selected query samples (8 samples per query setting) on the SYSU-MM01 dataset with our neural feature search method. Correct retrieved samples are in \textcolor{green}{green} boxes and wrong matchings are in \textcolor{red}{red} boxes (best viewed in color). Numerical values report cosine similarity scores of image pairs.} 
\label{fig:visualization}
\end{figure*}

As shown in Fig. \ref{fig:visualization}, the top-5 NFS retrieval results of 16 randomly selected query examples on the SYSU-MM01 dataset are plotted. We not only follow the original evaluation protocol, but also evaluate the \textit{Visible-Infrared} setting. 

In detail, the first column includes randomly selected samples from the query set, and retrieval results are sorted from left to right in descending order of cosine similarity scores. Due to lack of color information in IR images, some cases are even difficult for human (e.g., query D and K). But the proposed method can retrieve correct results, which demonstrates the effectiveness of NFS in narrowing the large modality gap. According to the retrieval results for query B, C, D, and I, we also observe that our method exhibits certain robustness for high sample noises such as background clutter and partial occlusions. Interestingly, we discover that even if some persons change their clothes (e.g., query E), NFS can still return accurate retrieval results by mining other discriminative cues, perhaps the face or shoes.  Another interesting phenomenon is that performance of \textit{Visible-Infrared} setting is usually better than that of \textit{Infrared-Visible} one. The main reason is that visible images often provide richer appearance information than their IR counterparts. Although there are still a few failure cases, most of these images (such as query F, O, and P) only present back views of the person-of-interest with limited identity-related information (e.g. face, texture, or logo of clothes). In conclusion, NFS exhibits promising performance in either query setting. 

\end{document}